\documentclass[10pt,twocolumn,letterpaper]{article}

\usepackage{iccv}
\usepackage{times}
\usepackage{epsfig}
\usepackage{graphicx}
\usepackage{amsmath}
\usepackage{amssymb}
\usepackage[accsupp]{axessibility}

\usepackage{tikz}
\usepackage{pgfplots}\pgfplotsset{compat=1.9}

\usepackage[pagebackref=true,breaklinks=true,letterpaper=true,colorlinks,bookmarks=false]{hyperref}

\iccvfinalcopy 

\def\iccvPaperID{27} 

\ificcvfinal\fi

\begin{document}

\newcommand{\bc}{\mathbf{c}}
\newcommand{\bd}{\mathbf{d}}
\newcommand{\bn}{\mathbf{n}}
\newcommand{\bp}{\mathbf{p}}
\newcommand{\bu}{\mathbf{u}}
\newcommand{\bv}{\mathbf{v}}
\newcommand{\bx}{\mathbf{x}}
\newcommand{\by}{\mathbf{y}}

\newcommand{\cC}{\mathcal{C}}

\newcommand{\argmin}{\operatornamewithlimits{arg\,min}}
\newcommand{\dist}{\operatorname{dist}}

\newcommand{\footremember}[2]{%
   \footnote{#2}
    \newcounter{#1}
    \setcounter{#1}{\value{footnote}}%
}
\newcommand{\footrecall}[1]{%
    \footnotemark[\value{#1}]%
} 

\title{Towards realistic symmetry-based completion of previously unseen point clouds}

\author{Taras Rumezhak${}^{1,2}$  \quad   
Oles Dobosevych${}^1$  \quad  
Rostyslav Hryniv${}^1$  \quad
Vladyslav Selotkin${}^2$ \\
Volodymyr Karpiv${}^2$  \quad
Mykola Maksymenko${}^2$ \\[5pt]
${}^1$The Machine Learning Lab at Ukrainian Catholic University \quad  ${}^2$SoftServe Inc. \\
\texttt{\small\{rumezhak, dobosevych, rhryniv\}@ucu.edu.ua} \quad \texttt{\small\{trume, vselo, vkarpi, mmaks\}@softserveinc.com}
}




\maketitle
\ificcvfinal\thispagestyle{empty}\fi


\begin{abstract}
3D scanning is a complex multistage process that generates a point cloud of an object typically containing damaged parts due to occlusions, reflections, shadows, scanner motion, specific properties of the object surface, imperfect reconstruction algorithms, etc. Point cloud completion is specifically designed to fill in the missing parts of the object and obtain its high-quality 3D representation. The existing completion approaches perform well on the academic datasets with a predefined set of object classes and very specific types of defects; however, their performance drops significantly in the real-world settings and degrades even further on previously unseen object classes.

We propose a novel framework that performs well on symmetric objects, which are ubiquitous in man-made environments. Unlike learning-based approaches, the proposed framework does not require training data and is capable of completing non-critical damages occurring in customer 3D scanning process using e.g. Kinect, time-of-flight, or structured light scanners. With thorough experiments, we demonstrate that the proposed framework achieves state-of-the-art efficiency in point cloud completion of real-world customer scans. We benchmark the framework performance on two types of datasets: properly augmented existing academic dataset and the actual 3D scans of various objects. The code is available here: \textcolor{red}{https://github.com/softserveinc-rnd/symmetry-3d-completion}
\end{abstract}

\section{Introduction}


\begin{figure}
\begin{center}
\includegraphics[width =\linewidth]{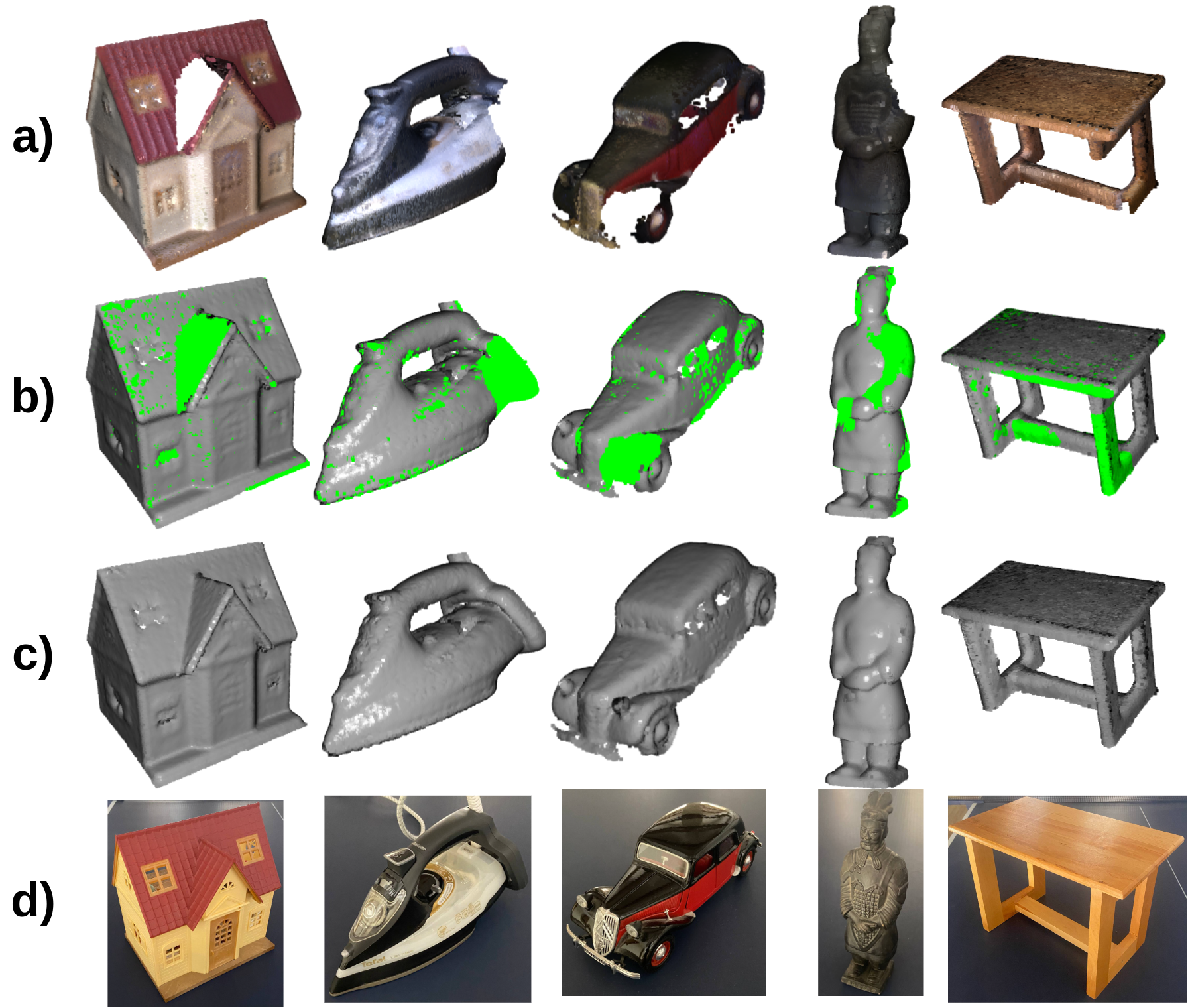}
\end{center}
  \caption{Method performance pipeline: a)~damaged object, b)~completed regions (green) c) result, d) original object}    
\label{fig:intro}
\end{figure}


The ability to recognize the geometry of a 3D object is a key pre-requisite in many  computer vision applications. This task can be accomplished with 3D scanners, including LIDARs, time-of-flight and structured light sensors that have seen tremendous progress over the last few years and have become more accessible and easy to use. 

Point cloud format is commonly used for the 3D shape representation of an object. Due to the limited sensor resolution and occlusions, an incomplete point cloud is typically generated as an output of a scanning pipeline. Recovering a damaged point cloud by filling in the missing parts, called \emph{point cloud completion}, is the  step required in many 3D scanning pipelines as it generates a more accurate 3D representation of the object and moreover enhances the overall performance of further 3D computer vision pipelines.

Most existing solutions of the point cloud completion task are based on deep neural networks trained on large datasets with a limited number of classes that the solution can complete. State-of-the-art approaches use datasets with e.g.\ up to 40 classes (ShapeNet \cite{shapenet2015}), which is a serious constraint for real-world applications as  reconstructing new types of objects requires a long training procedure and massive datasets that are hard to collect. 

To overcome this problem, we propose a novel framework that does not require training data and is capable of completing previously unseen symmetric objects. The symmetry of interest here is a mirror reflection symmetry that is present in most human artefacts. 

Our use case is different from those related e.g. to autonomous driving and for which some solutions have been proposed since they are mostly concentrated on completion of partially observed objects as in the KITTI dataset \cite{Geiger2013IJRR}. Our framework concentrates on a different use case with imperfect full object scans made with time-of-flight or structured light scanners, in which the damages can occur when the speed of scanner motion is too high or due to the shadows and occlusions. For example, with the True Depth camera, some regions can be missed when the speed of iPhone motion is greater than the point perception speed or when the object is on a turntable and images are taken by one camera. Another example is the structured light scanning pipeline where the projector as the source of light is set on the position appearing on one side of the scanning object and resulting shadows can lead to the damages in the resulting point cloud as seen in Figure~\ref{fig:intro}.

Our framework consists of three main steps. On the \emph{first step}, we approximate the symmetry plane of the given point cloud~$P$ using the Principal Component Analysis (PCA)~\cite{doi:10.1080/14786440109462720} applied to the surface normals in combination with a novel approach of Dominant convex hull directions. Then we construct the mirror reflection $P'$ of $P$ with respect to the approximated symmetry plane. Due to the broken symmetry in $P$ and imperfect symmetry plane approximation, $P'$ does not coincide with $P$. The \emph{second step} consists in the registration of the point clouds $P$ and $P'$ that would minimize the mismatch. We apply the Global registration algorithm of feature point-to-point matching via Fast Point Feature Histogram (FPFH)~\cite{5152473} along with Random Sample Consensus (RANSAC) with parameters suggested by Choi et al.~\cite{inproceedings}. The registration is iteratively improved with the Iterative Closest Point (ICP) algorithm~\cite{121791}.
On the \emph{third step}, we detect the missing parts of the point cloud~$P$ from its optimal mirror reflection~$P'$ using the proposed proximity measure and finish the completion of the point cloud~$P$ to better represent the initial 3D object. 

Performance of the proposed framework is demonstrated on point cloud completion of two datasets, including the standard academic dataset ShapeNet which was properly augmented for our use case with non-critical damage-rate as well as 3D scans of real-world objects collected for the purpose of this research. The accuracy of our algorithm was compared with several neural networks, such as  GRNet: Gridding Residual Network (GRNet)~\cite{xie2020grnet}, Morphing and Sampling Network (MSN)~\cite{liu2019morphing}, Point Completion Network (PCN)~\cite{yuan2019pcn}. In addition to not requiring any pre-training and being not restricted to particular object classes, our approach also achieves state-of-the-art completion results comparable to those obtained by other methods. Since the symmetry plane detection accuracy is critical for the overall performance of the framework, we also benchmark it on the 2017 ICCV Challenge: Detecting Symmetry in the Wild~\cite{8265408} dataset and report better or comparable results with the solution of Cicconet et al. \cite{8265416}, which was the only one with the provided code. Finally, we demonstrate that our framework generalizes to previously unseen real-world objects collected by three types of 3D scanners: Microsoft Kinect, iPhone’s TrueDepth camera (or LIDAR in the latest versions) and the structured light scanner.

The main contributions of the paper can be summarized as follows:
\begin{itemize}
    \item[-] a learning-free framework is proposed for point cloud completion based on the mirror symmetry;
    \item[-] a novel symmetry plane detection technique is introduced as an intermediate step of the framework;
    \item[-] a novel distance metric is developed to measure the accuracy of point cloud matching;
    \item[-] the experiments were conducted on the common academic dataset ShapeNet and on the newly collected real-world dataset of scans; the benchmarks indicate that the proposed framework performs well on the general class of symmetric objects and even outperforms state-of-the-art approaches on the provided datasets.
\end{itemize}

\section{Related work}

\subsection{Point cloud completion}
\textbf{Learning-based.} The pioneering approaches to point cloud completion are mainly focused on the 
usage of deep neural networks (NN). Several approaches are based on the multi-layer perceptron (MLP) due to its simplicity and solid representation power. Qi et al.~\cite{qi2017pointnet} introduce PointNet that is applied directly to a point cloud and correctly handles the permutation invariance of its input. At the same time, this class of methods aggregates features with symmetric functions, such as Max-pooling, which do not fully exploit the geometric structure of a point cloud. 
Xie et al.~\cite{xie2020grnet} solve the problem of the structural and contextual information loss by introducing 3D grids as intermediate representations as a regularization to unordered point clouds and based on such grids present the Gridding Residual Network (GRNet) which shows great results on the Completion3D benchmark \cite{authors}.  Also to overcome the structure information loss Groueix et al.~\cite{groueix2018atlasnet} introduced the method that represents a 3D shape as a collection of parametric surface elements and utilizes surface representation of the shape instead of voxel grids or point clouds. Similarly, Liu et al.~\cite{liu2019morphing} propose the method that also uses a collection of parametric surface elements. In the first stage they predict a complete coarse-grained point cloud and in the second stage the algorithm merges the predictions with the input point cloud with the proposed sampling method. In contrast to intermediate parametrization or voxelization, Yuan et al.~\cite{yuan2019pcn} propose a Point Completion Network (PCN) which operates directly on the point clouds without any structural or geometry assumption and completes them in a coarse-to-fine fashion.

\textbf{Geometry-based.} Besides learning-based approaches some attempts have been made with geometry-based methods which complete the shapes using geometry features without any external data. 
The geometry-based methods themselves can be categorized into surface interpolations and symmetry-driven methods. Sarkar et al.~\cite{sarkar2017learning} propose a 3D shape parameterization method by surface patches that performs interpolated shape completion of complicated surface textures. The results achieved by Sorkine et al.~\cite{1314506} used the connectivity of the mesh containing the geometric information that approximates a set of the control point in a least-squares manner and Zhao et al.~\cite{zhao_inproceedings} propose the method that can fill great holes in triangular mesh models by approximation of newly created normals.  We should note that the primary task of these papers is to fill small locally uncompleted regions by smooth interpolation and the information based on a mesh, which is not the use case with point clouds. Mitra et al.~\cite{mitra} discussed the strengths and limitations of existing 3D symmetry algorithms. Pauly et al.~\cite{pauly10.1145/1360612.1360642} introduced the framework for detecting the repeated geometric patterns in mesh-based models by the analysis of pairwise similarity transformations. Sung et al.~\cite{sung10.1145/2816795.2818094} method uses a 3D model collection to build the structural part-based priors to perform the completion. They also focus on the use case with low-quality consumer-level scanners and achieve compatible results, but still are the training-based approach.

\subsection{Symmetry plane detection}
Application of the symmetry plane detection for point clouds can be found in various fields of computer vision and computer graphics. Fan Xue et al.~\cite{fanxue} showed that symmetry estimation can be well accomplished with prior knowledge of the object type. However, in the general case, the object type is unknown beforehand. With an assumption of a bilateral object (having perfect or imperfect mirror symmetry) Combes et al.~\cite{combes} presented an iterative approach using the maximum likelihood principle and the expectation-maximization (EM) algorithm. To deal with the problem of noisy or missing data, Sipiran et al.~\cite{10.1111/cgf.12481} developed a feature-based approach. The algorithm generates a set of symmetry candidates based on detected features. The vote-based system validates all candidates and selects the best one. Several interesting methods were presented on the 2017 ICCV Challenge~\cite{8265408}: Detecting Symmetry in the Wild. Cicconet et al.~\cite{8265416} developed a simple and efficient symmetry estimation approach based on registration which is also a key step in our pipeline. However, the accuracy of the algorithm drops when the points in the symmetric parts are not perfectly aligned or missing parts are presented in the point clouds. Speciale et al.~\cite{PSpeciale2016ECCVfixed} propose the approach of detecting symmetries from incomplete data which then allows the scene completion. They extend standard symmetry detection techniques to exploit partially unexplored domains.
Cohen et al.~\cite{6247841} propose a method to recover symmetry relations using the geometry cues. The symmetry priors are incorporated in a new constrained bundle adjustment formulation. The method can perform 3D completion to get rid of Structure-from-Motion (SfM) artifacts based on the symmetries.

\section{Pipeline description}

Given a real-world 3D object $D$ that is symmetric with respect to a plane $\pi \subset \mathbb{R}^3$. The symmetry plane $\pi = \pi (\bp, \bn)$ is uniquely determined by any of its points $\bp\in\pi$ and a unit normal $\bn$; then the mirror reflection $T_\pi$ with respect to $\pi$ is given by \(T_\pi \bx = \bx - 2 \langle \bx - \bp, \bn \rangle\bn\), where $\langle \cdot,\cdot \rangle$ is the standard scalar product in $\mathbb{R}^3$. The object $D$ is called \emph{symmetric} with respect to $\pi$ if $T_\pi D = D$. 

In real-world applications, $D$ is represented by a point cloud \(P = \{\bx_1,\dots,\bx_n\}\) of points \(\bx_i \in D\). In the ideal situation, the points of $P$ are uniformly spread inside~$D$ so that the mirrored cloud~$T_\pi P$ is almost identical to $P$, in the sense that the Chamfer distance between $P$ and $T_\pi P$~\eqref{chamfer_distance} is small enough. However, in the realistic cases point clouds $P$ representing $D$ have holes, i.e., there are sub-regions $D_j \subset D$, $j=1,2, \dots, h$ with no or very few points $\bx_i$. Our goal is to fill in these holes, i.e., to find a point cloud $P^* \supset P$ that is symmetric with respect to a plane $\pi^*$ and has the smallest possible volume. In order that $P^*$ would represent the whole $D$, the missing part $\bigcup_{j=1}^h D_j$ should have a void (or negligibly small) intersection with its mirror image $T_\pi (\bigcup_{j=1}^h D_j)$. 

\begin{figure}
    \centering
    \includegraphics[width=0.4\textwidth]{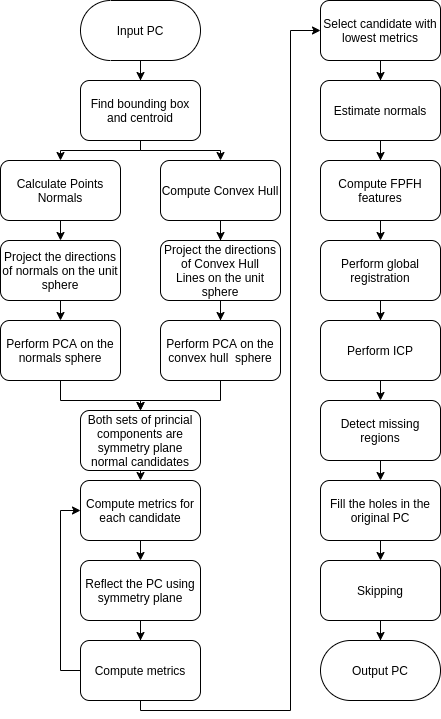}
    \caption{Pipeline of the algorithm}
    \label{fig:pipeline}
\end{figure}

The completion framework proposed in this paper (see Fig.~\ref{fig:pipeline}) includes three main steps: symmetry plane estimation, its iterative approximation, and then localization of the holes and their filling. The first two steps are based on point cloud registration and include several preliminary steps, such as bounding box and centroid computation, symmetry plane candidates computation and approximation.

\subsection{Bounding box and centroid computation}
Missing regions in the point cloud induce errors to the whole objects geometry estimation. To localize the point cloud and minimize the influence of its missing parts, we put it inside its \emph{bounding box}; this is a cuboid of minimal volume that is circumscribed around the point cloud and whose faces are parallel to the coordinate planes. It is constructed by detecting the extreme points of the point cloud along the coordinate $x$, $y$, and $z$-axes; the coordinates of the cuboid vertices are combinations of these extreme values in each direction:
\begin{align*}\label{bb_computation}
& \bv_{1} = (x_{\mathrm{min}}, y_{\mathrm{min}}, z_{\mathrm{min}}), \\
& \bv_{2} = (x_{\mathrm{max}}, y_{\mathrm{min}}, z_{\mathrm{min}}), \\
& \makebox[3.6cm]{\dotfill} \\ 
&  \bv_{8} = (x_{\mathrm{max}}, y_{\mathrm{max}}, z_{\mathrm{max}}). 
\end{align*}
The \emph{centroid} $\mathbf{c}$ of the bounding box has coordinates
 \begin{equation}\label{eq:centroid}
     \mathbf{c}= \Bigl( \frac{x_{\mathrm{max}}+x_{\mathrm{min}}}{2}, \frac{y_{\mathrm{max}}+y_{\mathrm{min}}}{2}, \frac{z_{\mathrm{max}}+z_{\mathrm{min}}}{2} \Bigr).
 \end{equation}

The bounding box centroid $\mathbf{c}$ is more stable to point cloud damages than the point cloud mass center
mass center \( 
     \overline{\bx} = \frac1{n} \sum_{i=1}^n \bx_i
\) 
and we prefer it as a reference point for the symmetry plane approximation in the dominant convex hull direction method of Subsection~\ref{ssec:dominant}.

\subsection{Symmetry plane estimation}
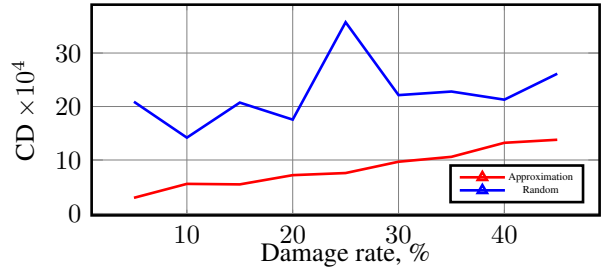
\begin{figure}[h]
\begin{tikzpicture}
  \begin{axis}[ 
  width=\linewidth,
  line width=1,
  grid=major, 
  tick label style={font=\normalsize},
  legend style={nodes={scale=0.4, transform shape}},
  label style={font=\normalsize},
  legend image post style={mark=triangle},
  legend pos = {south east},
  grid style={gray, very thin},
  xlabel={Damage rate, \%},
  ylabel={CD $\times 10^4$ },
   y tick label style={
    /pgf/number format/.cd,
    fixed,
    fixed zerofill,
    precision=0
    },
    yscale = .5 
  ]



    \addplot[red] coordinates
    {(5, 2.92813619)  (10, 5.53527373) (15, 5.44000013)  (20, 7.16478529)  (25, 7.56060472) (30, 9.67404271) (35, 10.59208765) (40, 13.21435619) (45, 13.78513004)};
    \addlegendentry{Approximation}
    
    \addplot[blue] coordinates
   {(5, 20.89) (10, 14.18) (15, 20.73) (20, 17.55) (25, 35.76) (30, 22.13) (35, 22.80) (40, 21.28) (45, 26.13)};
    \addlegendentry{Random}
  \end{axis}
\end{tikzpicture}
\caption{Comparison of the final results with a random initial symmetry plane approximation and suggested approximation using PCA candidates. The experiments were run on the properly augmented ShapeNet dataset with $9$ different damage rates}
\label{fig:pca_vs_random}
\end{figure}

Standard approaches to symmetry plane estimation fail to find the correct symmetry plane $\pi$ due to random point sampling and non-symmetric missing parts in the point cloud. However, the proposed algorithm requires only an approximation of the symmetry plane that is close enough to the correct one. We suggest two approaches to the symmetry plane estimation, which are based on the principal component analysis (PCA) applied to the surface normals and dominat convex hull directions, respectively. These two methods generate several candidates for the symmetry plane, and each candidate is evaluated with the balance distance metrics~\eqref{eq:BD} to select the best one. As experiments show, using this best candidate as initial symmetry plane approximation results in solutions with better Chamfer distance to the ground truth (see Fig.~\ref{fig:pca_vs_random}).



\subsubsection{Principal component analysis}
\label{ssec:PCA}
Principal component analysis (PCA) can be used to estimate the symmetry plane of a  point cloud $P$. The PCA calculates the eigenvalues and eigenvectors (principal components) of the $3\times 3$ covariance (i.e., inertia) matrix~$\Sigma$ of~$P$. 
If~$\pi$ is a symmetry plane of~$P$, then the principal components $\mathbf{c}$ are invariant under $T_\pi$, so that each principal component either belongs to $\pi$ or is orthogonal to it. 

Therefore, given a point cloud $P$ that is not completely symmetric with respect to $\pi$, we suggest a candidate $\pi^*$ for $\pi$ as a plane through the bounding box centroid~$\mathbf{c}$ of~\eqref{eq:centroid} that is orthogonal to a principal component. Assuming that $\Sigma$ has no repeated eigenvalues (which is the case when the point cloud has no multiple planes of symmetry), we thus get three candidates $\pi^*$ of which the best one is chosen.

Experiments showed that the PCA is not sufficiently robust to random damages; therefore, we apply it not to the initial point cloud~$P$ itself but to two point clouds that are projections onto the unit sphere~$S^2$ of the surface normals and the so called dominant convex hull directions discussed in Subsections~\ref{ssec:normal} and \ref{ssec:dominant}. 

\subsubsection{PCA of normal directions}\label{ssec:normal}

We use the \texttt{open3d.geometry.estimate\_normals} algorithm of Open3D\footnote[2]{http://www.open3d.org/} library~\cite{Zhou2018} to calculate the point cloud normals $\bn_j$ by locally fitting planes through the surface points of~$P$ and register them on the unit sphere~$S^2$. Then we run the PCA on the resulting set $N$ of normals to find a candidate $\pi^*$ of the symmetry plane~$\pi$ as explained in Subsection~\ref{ssec:PCA}. 

In the case where the point cloud $P$ was obtained from the uniformly representing point cloud of a symmetric object~$D$ by removing several randomly located holes, the PCA of normal directions in most cases returns a good approximation $\pi^*$ for the ground-truth symmetry plane $\pi$.

\subsubsection{Dominant convex hull directions}\label{ssec:dominant}

Our experiments demonstrated that there are cases when the PCA of normal directions fails to suggest a good candidate for the symmetry plane~$\pi$. A reason for that is that the missing parts $D_j$ of the point cloud~$P$ may significantly  affect the surface and thus the normals. To overcome this problem, we propose a complementary approach for symmetry plane estimation using the PCA of the dominant directions in the point cloud convex hull. 

Firstly, the convex hull is built around the point cloud~$P$ using the Quickhull algorithm~\cite{10.1145/235815.235821} implemented in Open3D library. It consists of numerous intervals joining points on the point cloud surface, and we register their directions $\pm\bu_j$ on the unit sphere $S^2$. If the point cloud $P$ has mirror symmetry~$\pi$, then the set of unit directions $\bu_j$ will be almost  symmetric with respect to~$\pi$.

\begin{figure}[ht]
\begin{center}
\includegraphics[width= 0.9\linewidth]{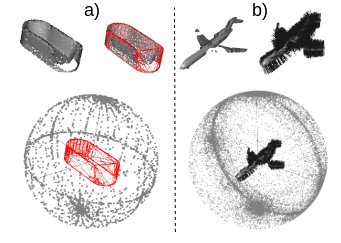}
\end{center}
   \caption{ a) Dominant convex hull directions: Projections of the convex hull lines onto the unit sphere $S^2$; 
   b) Projections of the points normals onto the unit sphere $S^2$}
\label{fig:sphere}
\end{figure}

In presence of measurement errors and missing parts $D_j$, the point cloud~$P$ will not be symmetric with respect to~$\pi$ but the set $U$ of unit directions on the sphere~$S^2$ is typically close to symmetric, see Figure~\ref{fig:sphere}. 
We apply the PCA to the set $U$ of directions on the sphere $S^2$ and get three candidates for a symmetry plane as  described in Subsection~\ref{ssec:PCA}.

\subsection{Metric}

To measure the symmetry plane estimation accuracy and to choose the best among several candidates, a new metric is suggested.
For a point cloud~$P$ and a symmetry plane candidate $\sigma$, we first calculate the mirror symmetric image~$P' = T_\sigma(P)$ of~$P$. Denote by $Q(\bx, d)$ the cube of center $\bx$ and side length~$d$, whose faces are parallel to the coordinate planes; we call a point $\bx \in P \cup P'$ \emph{balanced} if 
\begin{equation}\label{eq:balanced_point}
    \frac{\bigl| |Q(\bx,d)\cap P| - |Q(\bx,d)\cap P'|\bigr|}{|Q(\bx,d)\cap P| + |Q(\bx,d)\cap P'|} \le \varepsilon
\end{equation}
for experimentally established threshold $\varepsilon$; here $|A|$ denotes the cardinality of a set $A$. Denote by $B(P,P')$ the number of all balanced points in $P \cup P'$; then the \emph{balanced distance} $BD (P,P')$ between the point clouds $P$ and $P'$ is 
\begin{equation}\label{eq:BD}
    BD(P,P')= 1- \frac{B(P,P')}{|P|+ |P'|};
\end{equation}
this is an analogue of the standard \emph{intersection-over-union} measure.

\subsection{Registration}
    The registration of the point cloud is required since the candidate for the symmetry plane is only an estimation.  The common method for point cloud alignment is Iterative Closest  Point  (ICP), which is known as the local registration method because it relies on a rough initial alignment. In contrast, we will use global registration which does not require the initial alignment but produces a less accurate result, so it will be used as the input of ICP which will perform the post-processing refinement.

\subsubsection{Normals and FPFH features}
    The FPFH features~\cite{5152473} describe the local geometric properties of a point cloud and are constructed of the vectors with 33-dimensions. 
    Computation of the FPFH features requires point normal estimation, which is based on a local neighborhood of the point and involves covariance analysis. The camera position or the alignment axis should be provided to choose one of the two directions consistently.  
    
\subsubsection{Global registration}
    Extracted FPFH features can be used to perform the registration of the point clouds. The registration based on the RANSAC algorithm with parameters proposed by Choi et al.~\cite{inproceedings} is slow due to numerous model proposals and evaluations, so it is performed on the down-sampled point cloud. The Fast Global Registration~\cite{fast_global}  algorithm optimizes the process of finding point correspondences as for each iteration it has no model proposal and evaluation involved.

\subsubsection{Refine registration via ICP}
Iterative Closest Point~\cite{121791} is a classical approach for point cloud registration. The algorithm finds the best rotation matrix $R$ and translation vector $\mathbf{t}$ to minimize the squared error defined as
     \begin{equation}\label{icp_computation}
    E(R, t) = \frac{1}{N_{p}}\sum_{i=1}^{N_{p}}\left | \left | \mathbf{x}_{i}-R\mathbf{p}_{i} - \mathbf{t}\right | \right |^{2},
    \end{equation}
    where \(\mathbf{x}_{i}\) and \(\mathbf{p}_{i}\) are the corresponding points in point clouds $P$ and $P'$ to be aligned, and \(N_{p}\) is the number of corresponding point pairs. 
    
 If the correct correspondences are known, then the alignment can be performed accurately in one step. However, usually, the matching points are not known beforehand, and the ICP finds the optimal transformations iteratively.
    
\subsection{Hole detection and filling}

After the best point cloud alignment of $P$ and its mirrored image $P'$ has been achieved (i.e., the best symmetry plane $\pi^*$ proposed), the hole detection algorithm is applied. Every point $\bx'$ in the mirrored point cloud $P' = T_{\pi^*}(P)$ that is not balanced in the sense of~\eqref{eq:balanced_point} is in a hole of $D$ if $|Q(\bx',d) \cap P'|$ is bigger than $|Q(\bx',d) \cap P|$. The holes in the original point cloud are detected this way along with the corresponding parts of the mirrored point cloud $P'$ that should be added to the point cloud $P$ to finish completion.
 

\subsection{Skipping}\label{ssec:skipping}

Performance analysis of the proposed pipeline showed that there are some cases where the completed point clouds have significantly bigger Chamfer distance~\eqref{chamfer_distance} to the ground truth than the original one. One of the main reasons for this is that not all objects in the datasets have mirror symmetry; in some other cases the algorithm fails to suggest a good symmetry plane approximation. 

To deal with this issue, we add a validation step that decides whether the completed point cloud $P^*$ or the original damaged point cloud $P$ should be returned. After the completion, we calculate the scaled Chamfer distance $CD(P, P^*)/s$ between~$P$ and $P^*$; if it exceeds a threshold value~$d^*$, then we expect that the Chamfer distance between the completion~$P^*$ and the ground truth will be large; we thus regard~$P^*$ as a poor reconstruction and return the original~$P$ instead. This step lets us judge whether the object is symmetric with non-critical damages.

The threshold value $d^*$ was chosen experimentally based on the corresponding performance curve of~$d^*$. The scale~$s$ is chosen as the average distance between the closest points in the input point cloud~$P$. 
\graphicspath{ {./graphics/dataset/} }



\section{Experiments}
For the evaluation purpose, the ShapeNet dataset \cite{shapenet2015} was chosen as it is the most frequently used source for 3D completion applications. We first explain the way how  evaluation dataset was formed, then describe the metrics used for performance evaluation of the proposed algorithm and comparison with other methods on point cloud completion and mirror symmetry detection. We then report the corresponding scores for both tasks and, finally, show point cloud completion performance of the proposed method on real-world 3D object scans in the wild.

\begin{figure}[t]
\begin{center}
\includegraphics[width=\linewidth]{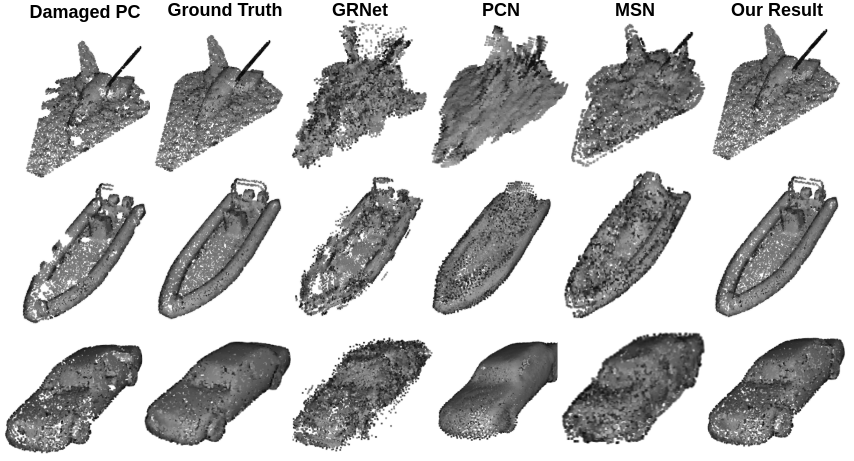}
\end{center}
  \caption{Comparison of methods on damaged ShapeNet objects.}
\label{fig:damages}
\end{figure}

\begin{figure}[t]
\begin{center}
\includegraphics[width=\linewidth]{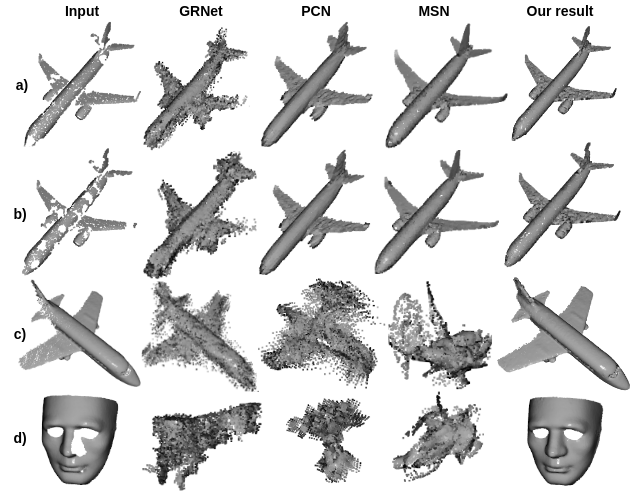}
\end{center}
  \caption{Comparison of methods on synthetic data with: a) 15\%, b) 35\% damage rate, and real scans with classes on which compared approaches were: c) trained on, d) not trained on.}
\label{fig:damages2}
\end{figure}


\begin{figure}[h]
\begin{tikzpicture}
  \begin{axis}[ 
  width=\linewidth,
  line width=1,
  grid=major, 
  tick label style={font=\normalsize},
  legend style={nodes={scale=0.4, transform shape}},
  label style={font=\normalsize},
  legend image post style={mark=triangle},
  legend pos = {south east},
  grid style={gray,very thin},
  xlabel={Damage rate, \%},
  ylabel={CD $\times 10^4$ },
  y tick label style={
    /pgf/number format/.cd,
    fixed,
    fixed zerofill,
    precision=0
    },
    yscale = .75 
 ]

    \addplot[red] coordinates
      {(5, 2.92)  (10, 5.53) (15, 5.44)  (20, 7.16)  (25, 7.56) (30, 9.67) (35, 10.59) (40, 13.21) (45, 13.79)};
      \addlegendentry{Ours}

    \addplot[blue] coordinates
      {(5, 8.95) (10, 9.02) (15, 8.82) (20, 8.88) (25, 8.98) (30, 9.02) (35, 9.16) (40, 9.21) (45, 9.11)};
     \addlegendentry{GRNet}
     
     \addplot[green] coordinates
      {(5, 10.63) (10, 10.64) (15, 10.60) (20, 10.68) (25, 10.61) (30, 10.67) (35, 10.57) (40, 10.59) (45, 10.70)};
     \addlegendentry{PCN}
     
     \addplot[yellow] coordinates
      {(5, 11.32) (10, 11.35) (15, 11.17) (20, 11.12) (25, 11.23) (30, 11.18) (35, 11.37) (40, 11.56) (45, 11.42)};
     \addlegendentry{MSN}
     
  \end{axis}
\end{tikzpicture}
\caption{Comparison of methods on the ShapeNet based dataset with different damage rates.}
\label{fig:comparison_shapenet}
\end{figure}
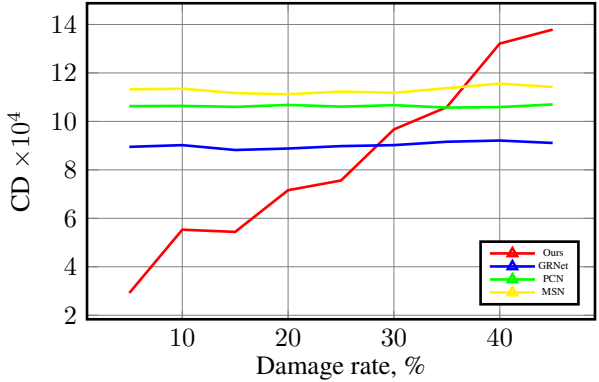


\begin{table}
\begin{center}
\begin{tabular}{|l|ccccc|}
\hline
Method & 5\% & 15\% & 25\% & 35\% & 45\% \\
\hline\hline
MSN & 11.32 & 11.16 & 11.23 & 11.37 & 11.42 \\
PCN & 10.63 & 10.60 & 10.61 & 10.57 & 10.70 \\
GRNet & 8.95 & 8.83 & 8.98 & \textbf{9.16} & \textbf{9.12} \\
Ours (-Skip) & 3.27 & 8.43 & 15.43 & 19.42 & 25.67 \\
Ours & \textbf{2.92} & \textbf{5.44} & \textbf{7.56} & 10.59 & 13.78\\
\hline
\end{tabular}
\end{center}
\caption{Numerical Chamfer Distance comparison based on different damage rate percentage. (-Skip) - without Skipping step.}
\end{table}

\subsection{Datasets creation} \label{datasets}

Performance evaluation is measured on an appropriately augmented ShapeNet dataset which is composed of 8 categories with $30\,974$ 3D models already uniformly sampled with $16\,384$ points on the mesh. In some popular benchmarks (e.g.,  Completion3D \cite{topnet2019}) the mentioned dataset was damaged sufficiently with a damaged rate (a ratio of a missing surface) up to 70\%, which is a normal case in autonomous driving when LIDAR captures only one side of the car and the algorithm must complete the whole unseen region. As our case is different, the type of damages must also differ from the existing benchmark. We expect to have the objects with 15-30\% damage rate, which is natural for our case. 
To perform experiments of various difficulty, we created nine versions of dataset with the damage rate $DR$ from $5\%$ to $45\%$ with the step of $5\%$ and a random number of missing regions fluctuating between $\lfloor 0.7DR \rfloor$ and $\lceil 0.95DR \rceil$. The random number of damaged regions, their size and location is a great test for the robustness of our algorithm. In learning-based approaches, robustness could be achieved by noise-based data augmentation, while in our case random damaged regions are the inverse of the noise. 





\subsubsection{Real-world data}
To assess framework performance on real data, we also collected a dataset of more than 200 scanned objects. Large objects such as a chair were scanned with the Microsoft Kinect v2, while small objects such as a toy plane were scanned with the iPhone’s TrueDepth camera (or LIDAR) or our structured light 3D scanner.

Learning-based approaches were trained on the academic dataset and do not generalize well to the real data. To fit the input standard of compared approaches, the obtained objects were downsampled, properly scaled and translated accordingly to contain the fixed number of points similar to the original ShapeNet dataset. Even on the categories the compared approaches were trained on, their performance remains poor. While our framework is not limited to a given type and size of point clouds or a defect so it successfully completes 3D scans of real objects as shown in the Fig.~\ref{fig:damages2}.

\begin{table*}[t] 
\begin{center}
\begin{tabular}{|l|ccccc|}
\hline
Method & Original & Damaged by 10\% & Damaged by 20\% & Damaged by 30\% & Damaged by 40\% \\
\hline\hline
Symmetry via Registration & 68.63\% & 45.75\% & 44.44\% & 42.44\% & 35.29\%\\
Ours & 66.24\% & 50.32\% & 49.67\% & 48.36\% & 45.09\%\\
\hline
\end{tabular}
\end{center}
\caption{Accuracy comparison with Symmetry via Registration with threshold $\theta = 0.2$}
\label{tab:symmetry_comparison_table}
\end{table*}

\subsubsection{Symmetry Dataset}

One of the key steps of our framework is the estimation of the best symmetry plane. To evaluate the performance of the proposed algorithm on this step, we benchmarked it on the 3D Global Reflection Symmetry dataset from 2017 ICCV Challenge: Detecting Symmetry in the Wild~\cite{8265408} consisting of $1\,795$ models with up to three symmetry planes in each. 




\subsection{Evaluation metrics}
\subsubsection{Chamfer distance}

As a proximity measure between two point clouds $P_1$ and $P_2$, we use the Chamfer distance~\cite{fan2016point, 8953650} that calculates the average closest point distance: 
\begin{equation}\label{chamfer_distance}
\begin{aligned}
CD(P_{1}, P_{2}) &=  \frac{1}{\left|P_{1}\right|} \sum_{\bx \in P_{1}} \dist(\bx, P_{2}) \\ &+ \frac{1}{\left|P_{2}\right|} \sum_{\by \in P_{2}} \dist(\by, P_1)
\end{aligned}
\end{equation} 
\subsubsection{Symmetry Plane Detection Accuracy}
 
As proposed in the ICCV 17 Challenge~\cite{8265408}, the 3D planar reflective symmetries are evaluated according to the position and orientation of the symmetry plane with respect to the ground truth. The detected symmetry is considered correct if both the position of the reference point is on the plane and the normal is close to ground truth with some threshold. The symmetry is rejected if the cosine distance between the symmetry normals is above a threshold: $\arccos \left(\left|n \cdot n_{\mathrm{GT}}\right|\right)>\theta$. 
A symmetry is also rejected if the distance of the tested center $\bc$ to the ground truth plane restricted to the bounding parallelogram is larger than a threshold $\left\|\bc-\Pi_{\mathrm{GT}}(\bc)\right\|_{2}>\tau$.
Accuracy is the relative frequency of correct predictions among all objects.

\subsection{Comparison with other methods}
\subsubsection{Point Cloud Completion}

We evaluate the completion results on the same datasets subject to different damage rates. GRNet, the top method of the Completion3D benchmark, shows good results in general, but approximately $17\%$ of its outputs have a very large Chamfer distance thus leading to a moderate average metric. MSN gives worse results than GRNet and still shows worse performance on small damages than our method. Our framework has an $8\%$ of anomaly big metric values due to a poor symmetry plane detection or its absence in the original object; applying the Skipping validation, we significantly increase the average results.

Weak sensitivity of GRNet, PCN and MSN to damage rate can be explained by the fact that the corresponding neural networks are trained on particular classes of the Completion 3D benchmark dataset (based on ShapeNet) and thus can complete point clouds of the learned objects with even 70-80\% of damage. Our method, on the contrary, would not be able to successfully reconstruct so heavily damaged object; however, scans of so poor quality are highly unlikely to be produced in a multiview scanning. Even without Skipping validation, our results are comparable with state-of-the-art approaches on objects with at most $20\%$ damage rate, which is quite natural for the real world, while the final result is better with up to $30\%$ damage rate and degrades only after $35-40\%$. PCN and GRNet results are practically the same, see Fig.~\ref{fig:comparison_shapenet}, however, PCN overall accuracy is lower. Also, see the visual comparison on the damaged ShapeNet objects in Fig.~\ref{fig:damages}.




\subsubsection{Symmetry Plane Detection Accuracy}

We have compared the achieved symmetry plane estimation accuracy to the Symmetry via Registration (SvR) algorithm~\cite{8265416}. SvR is based on the optimal symmetric pairwise assignment of curves. This algorithm outperforms ours on the non-damaged objects by $2.4\%$. However, its quality significantly deteriorates with damage rate growth, while the proposed method gives more stable results; see Tab.~\ref{tab:symmetry_comparison_table}.


\section{Limitations}
There are limitations in our approach: non-symmetric objects, symmetric damages, and their sizes. We need to notice that in the case of trained models non-symmetric objects could be addressed. However, each such non-symmetric class has to be introduced separately, while our method is readily applicable to new symmetric objects. Symmetric damages with a damage rate up to 30\% are not likely, but in our case, especially in the structured light pipeline, such damages are highly unlikely as shadows or glares can appear only on one side of the object.
For our use case, we focus on the moderate damage rate of 15 to 30\% which is normal for multiview structured light or hand SfM scanners. Another must mention point is the time performance of our method. For the 16K points ShapeNet model it takes 1.3 seconds on the Python CPU implementation to get the result while the compared approaches do not support CPU.
\section{Conclusion}


Point cloud completion is a critical step in real-world 3D scanning pipelines where objects are typically damaged due to the imperfections of the scanning process or reconstruction algorithms. Being an active field of research, the latest SOTA results are often associated with trained NN-based models that learn peculiarities of different classes of objects. While for specific cases such approaches showcase impressive results in the general case of new types or unseen objects the geometry-based methods can show more robust performance.

In this paper we present a novel point cloud completion framework that exploits symmetries of the objects to fill in the damaged parts of the objects. The method requires no training data and yet provides better or comparable performance than recent SOTA NN-based approaches such as PCN, MSN or GRNet. 

\emph{First}, we explore the robustness of our algorithm on academic benchmark datasets that were used to train the above models. Here for low damage rates below 30\% our method clearly outperforms other approaches and showcases Chamfer distance of 3 to 9. For the higher damage rates up to 45\%, it shows comparable to SOTA performance. \emph{Next}, we present a new real-world dataset of more than 200 objects and showcase qualitatively that our approach is robust to variation in object shape, while trained methods in some cases fail on previously unseen data (see Fig.~\ref{fig:damages2}). The robustness of our pipeline is linked to a novel method of symmetry plane approximation that we outline and benchmark in the manuscript. In comparison to the benchmark SvR algorithm it provides a significant boost in performance on damaged point clouds as showcased in Table~\ref{tab:symmetry_comparison_table}. \emph{Finally}, to reproduce our results, we provide the code of algorithm implementation. 

As the proposed Point Cloud completion pipeline was integrated and tested in a production 3D scanning setting, we believe that it has a wide potential as a robust engineering solution for a wide range of practical applications. 

{\small
\bibliographystyle{ieee_fullname}
\bibliography{egbib}
}

\end{document}